\documentclass[runningheads]{llncs}
\input{psfig.sty}

\usepackage{epsfig}
\usepackage{graphics}

\usepackage{amssymb}

\usepackage[latin1]{inputenc} % inclui acentuação latina

\begin{document}

\pagestyle{headings}
%In order to omit page numbers and running heads
%please change this line to
%\pagestyle{empty}
%and change the first command line too, see above.

\mainmatter

\title{Face Verification in Polar Frequency Domain: a Biologically Motivated Approach}
\titlerunning{Face Verification in Polar Frequency Domain}

\author{Yossi Zana\inst{1} \and Roberto M. Cesar-Jr\inst{1} \and Rogerio S. Feris\inst{2} \and Matthew Turk\inst{2}
\thanks{This work was supported by FAPESP (03/07519-0, 99/12765-2) and CNPq (150409/03-3).
}
}                     % Do not remove
\institute{Dept. of Computer Science, IME-USP, Brazil, \{zana,cesar\}@vision.ime.usp.br \and University of California, Santa Barbara, \{rferis,mturk\}@cs.ucsb.edu}

\maketitle

% \vspace*{-0.5cm}

\begin{abstract}
We present a novel local-based face verification system whose components are analogous to those of biological systems. In the proposed system, after global registration and normalization, three eye regions are converted from the spatial to polar frequency domain by a Fourier-Bessel Transform. The resulting representations are embedded in a dissimilarity space, where each image is represented by its distance to all the other images. In this dissimilarity space a Pseudo-Fisher discriminator is built. ROC and equal error rate verification test results on the FERET database showed that the system performed at least as state-of-the-art methods and better than a system based on polar Fourier features. The local-based system is especially robust to facial expression and age variations, but sensitive to registration errors.
\end{abstract}

\section{Introduction}
Face verification and recognition tasks are highly complex task due to the many possible variations of the same subject in different conditions, like facial expressions and age. Most of the current face recognition and verification algorithms are based on feature extraction from a Cartesian perspective, typical to most analog and digital imaging systems. The human visual system (HVS), on the other hand, is known to process visual stimuli by fundamental shapes defined in polar coordinates, and to use logarithmical mapping. In the early stages the visual image is filtered by neurons tuned to specific spatial frequencies and location in a linear manner \cite{DevaloisDevalois1990}. In further stages, these neurons output is processed to extract global and more complex shape information, such as faces \cite{Perretetal1982}. Eletrophysiological experiments in monkey's visual cerebral areas showed that the fundamental patterns for global shape analysis are defined in polar and hyperbolic coordinates \cite{Gallantetal1993}. Global pooling of orientation information was also showed by psychophysical experiments to be responsible for the detection of angular and radial Glass dot patterns \cite{WilsonWilkinson1998}. Thus, it is evident that information regarding the global polar content of images is effectively extracted by and is available to the HVS. Further evidence in favor of a polar representation use by the HVS is the log-polar manner in which the retinal image is mapped onto the visual cortex area \cite{Schwartz1977}. An analogous spatial log-polar mapping was explored for face recognition \cite{TistarelliGrosso1998}. One of the disadvantages of this feature extraction method is the rough representation of peripheral regions. The HVS compensates this effect by eye saccades, moving the fovea from one point to the other in the scene. Similar approach was adopted by the face recognition method of \cite{TistarelliGrosso1998}. 

An alternative representation in the polar frequency domain is the 2D Fourier-Bessel transformation (FBT)  \cite{Bowman1958}. This transform found several applications in analyzing patterns in a circular domain \cite{Foxetal2003}, but was seldom exploited for image recognition. In \cite{ZanaCesar2005} we suggested the use of global FB descriptors for face recognition algorithms. The present paper is a major development of this idea. The main contribution of the current work is the presentation and exhaustive evaluation of a face verification system based of local, in contrast to global extraction of FB features. Results show that such a system achieve state-of-the-art performance on large scale databases and significant robustness to expression and age variations. Moreover, we automated the face and eyes detection stage to reduce dependency on ground-truth information availability.

The paper is organized as follows: in the next two sections we describe the FBT and the proposed system. The face database and testing methods are introduced in Section 5. The experimental results are presented in Section 6 and in the last section we discuss the results.

\section{Polar Frequency Analysis}
The FB series \cite{Foxetal2003} is useful to describe the radial and
angular components in images. FBT analysis starts by converting the
coordinates of a region of interest from Cartesian $\left( {x,y}
\right)$ to polar $\left( {r,\theta } \right)$. The $f\left( {r,\theta } \right)$ function is represented by the
two-dimensional FB series, defined as

\begin{equation}
\label{eq:ifbt}
f(r,\theta )=\sum\limits_{i=1}^\infty {\sum\limits_{n=0}^\infty
  {A_{n,i}Jn(\alpha _{n,i}r)\cos (n\theta )} }
+\sum\limits_{i=1}^\infty {\sum\limits_{n=0}^\infty {Bn,iJn(\alpha
_{n,i}r)\sin (n\theta )} }
\end{equation}
where $J_n$ is the Bessel function of order $n$, $f(R,\theta )=0$
and $0\le r\le R. \quad \alpha _{n,i}$ is the $i{th}$ root of the
$J_n$ function, i.e. the zero crossing value satisfying
$J_n(\alpha _{n,i})=0$ is the radial distance to the edge of the
image. The orthogonal coefficients $A_{n,i}$ and $B_{n,i}$ are
given by

\begin{equation}
A_{0,i}=\frac{1}{\pi R^2J^2_1(\alpha _{n,i})}\int\limits_{\theta
  =0}^{\theta =2\pi
} {\int\limits_{r=0}^{r=R} {f(r,\theta )rJ_n(\frac{\alpha n,i}{R}r)drd\theta 
} } 
\end{equation}
if $B_{0,i}=0$ and $n=0$;

\begin{equation}
\left[ {\begin{array}{l}
      A_{n,i} \\
      B_{n,i} \\
        \end{array}} \right]=\frac{2}{\pi R^2J^2_{n+1}(\alpha _{n,i})}
    \int\limits_{\theta =0}^{\theta =2\pi } {\int\limits_{r=0}^{r=R} {f(r,\theta     )rJ_n(\frac{\alpha _{n,i}}{R}r) 
\left[ {\begin{array}{l}
           \cos (n\theta ) \\
           \sin (n\theta ) \\
        \end{array}} \right]drd\theta
	} }
\end{equation}
if $n>0.$

However, polar frequency analysis can be done using other
transformations. An alternative method is to represent images by
polar Fourier transform descriptors. The polar Fourier transform is
a well known mathematical operation where, after converting the
image coordinates from Cartesian to polar, as described above, a
conventional Fourier transformation is applied. These descriptors
are directly related to radial and angular components, but are not
identical to the coefficients extracted by the FBT.

\section{The algorithm}
The proposed algorithm is based on two sequential steps of feature extractions, and one classifier building. First we extract the FB coefficients from the images. Next, we compute the Cartesian distance between all the FBT-representations and re-define each object by its distance to all other objects. In the last stage we train a pseudo Fisher classifier. We tested this algorithm on the whole image (global) or the combination of three facial regions (local).

\subsection{Spatial to polar frequency domain}
\begin{figure} 
  \begin{center}
  \begin{tabular}{ccc}
     
      Gallery &
      Expression &
      Age \\
      \epsfig{file=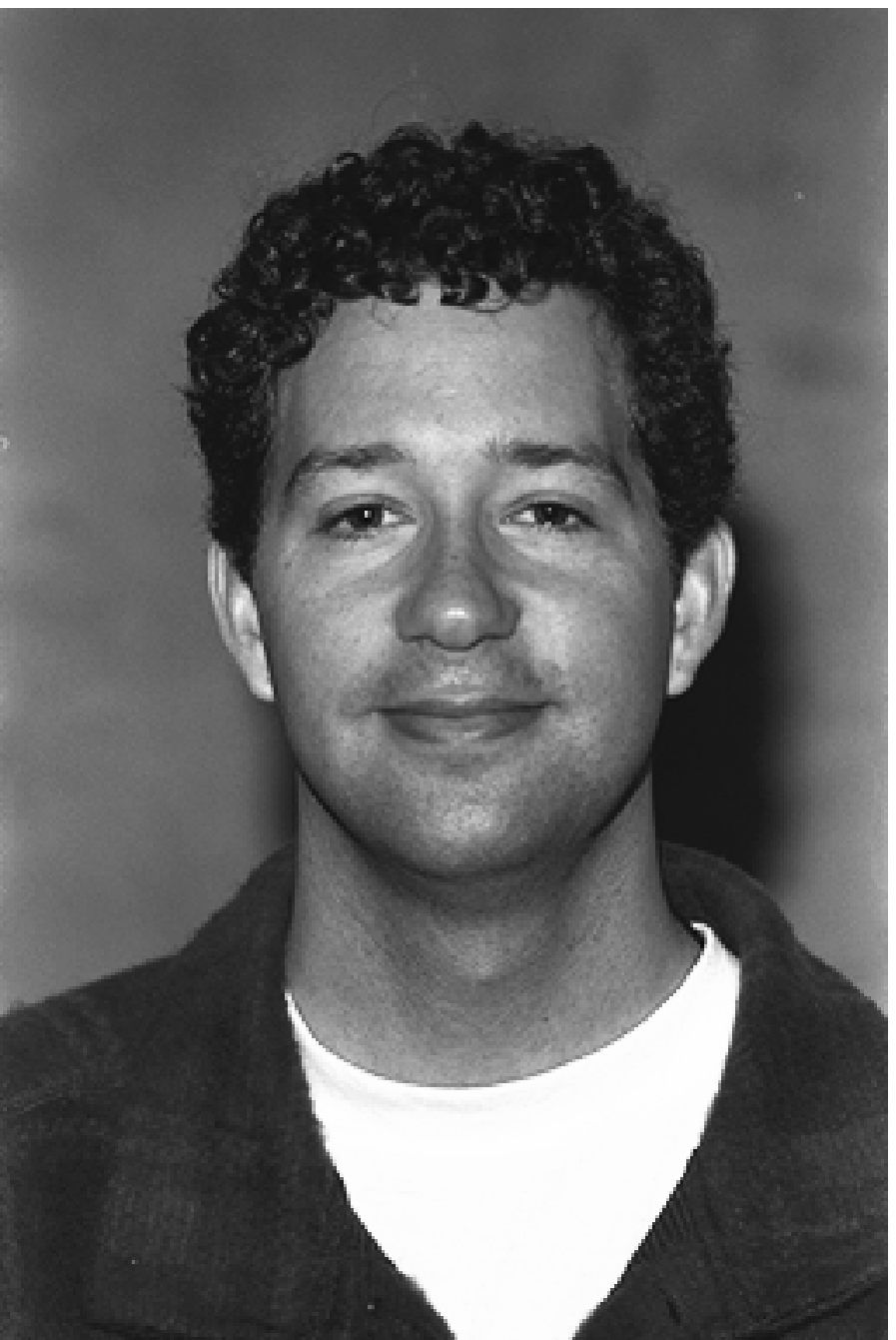, width=0.1\textwidth} &
      \epsfig{file=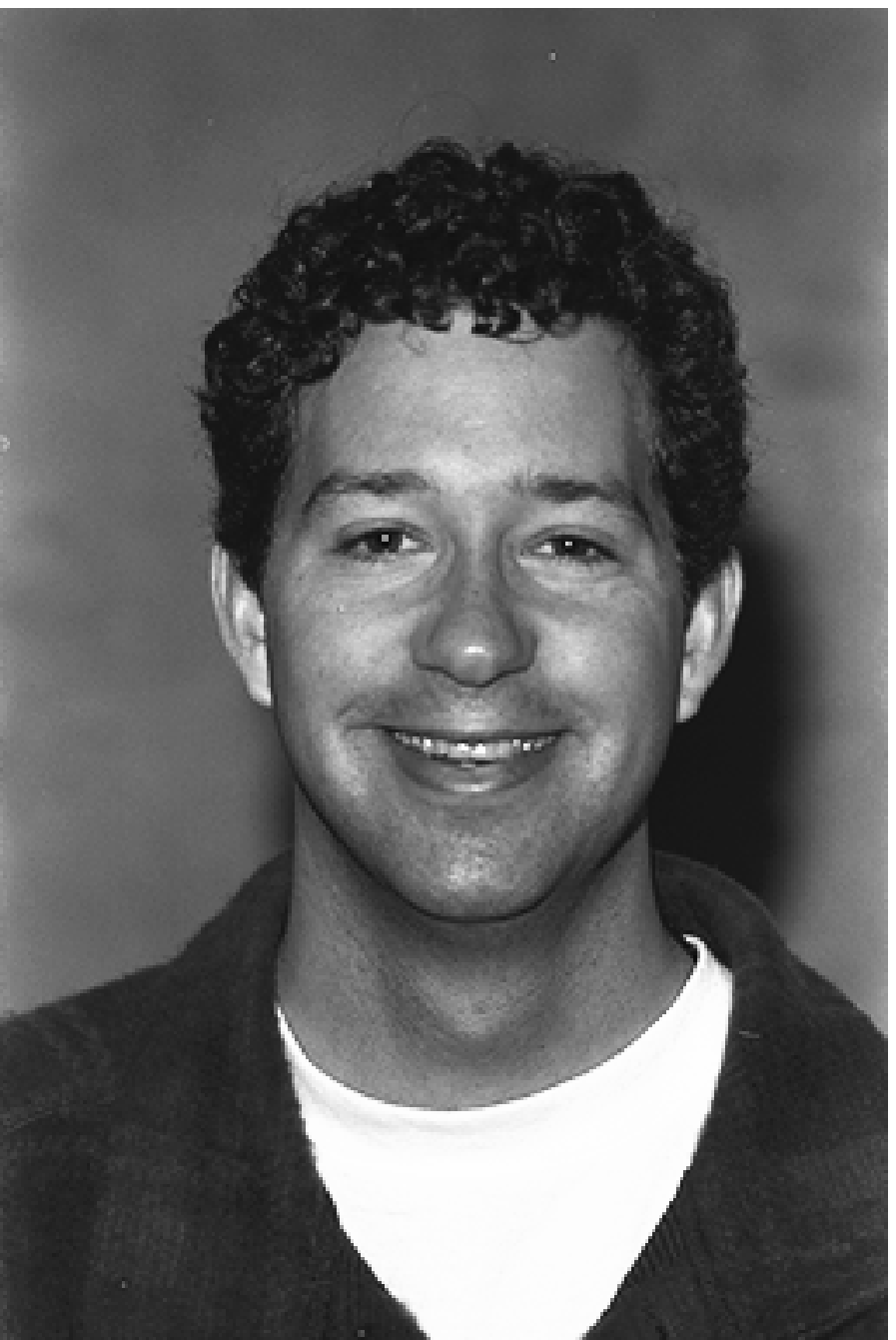, width=0.1\textwidth} &
      \epsfig{file=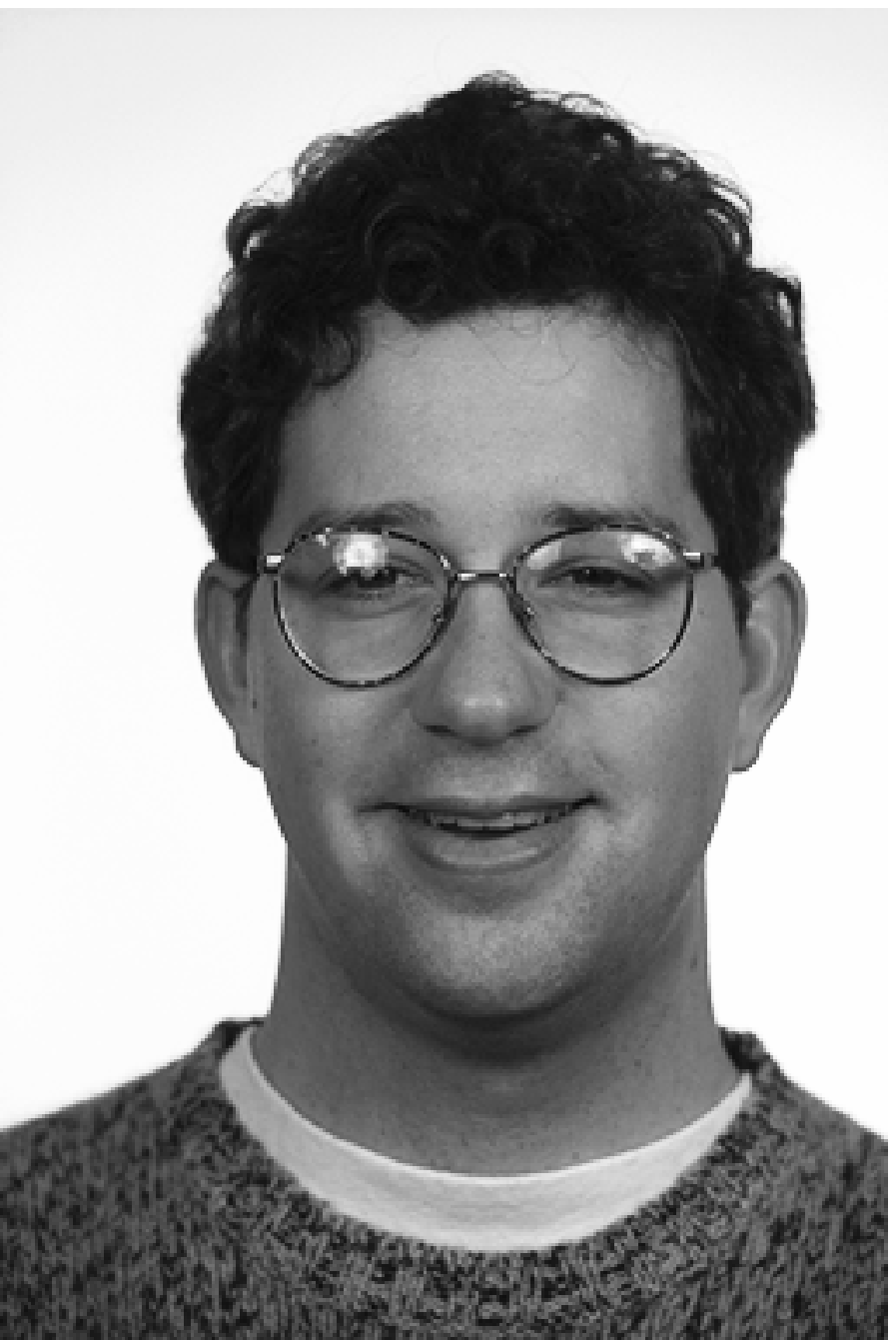, width=0.1\textwidth}
      \\
      Normalized &
      &
      Inverse FBT \\
      \epsfig{file=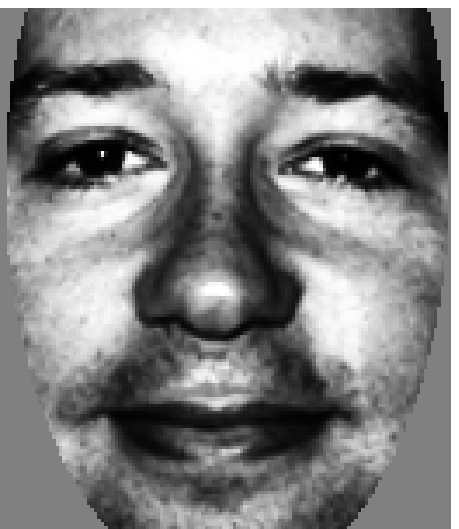, width=0.1\textwidth} &
      &
      \epsfig{file=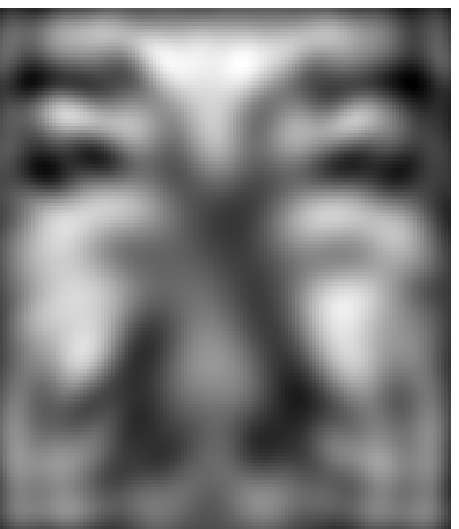, width=0.1\textwidth}
      \\
      \\
      \epsfig{file=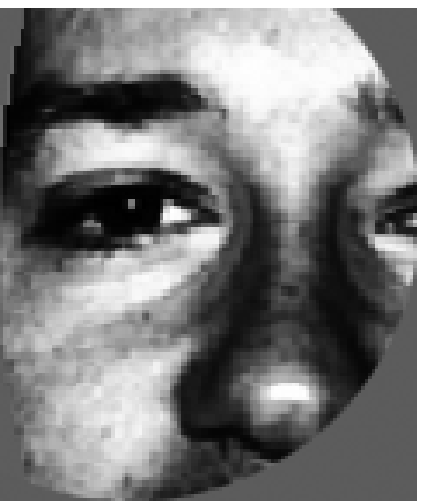, width=0.1\textwidth} &
      \epsfig{file=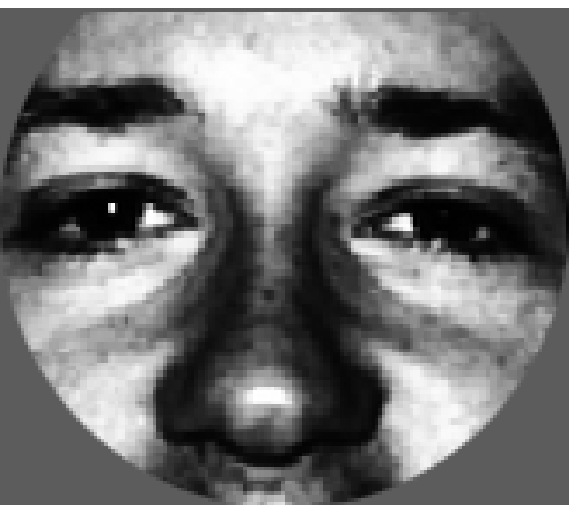, width=0.14\textwidth} &
      \epsfig{file=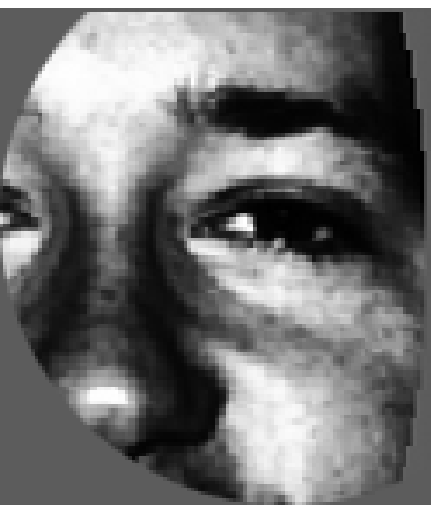, width=0.1\textwidth}

    \end{tabular}
  \caption{1$^{st}$ row: Samples from the datasets. 2$^{nd}$ row: Normalized whole face and the FB inverse transformation. 3$^{rd}$ row: The regions that were used for the local analysis}
  \label{fig:example}
  \end{center}
\end{figure}

Images were transformed by a FBT up to the 30$^{th}$ Bessel order and 6$^{th}$ root with angular resolution of 3\r{ }, thus obtaining to 372 coefficients. These coefficients correspond to a frequency range of up to 30 and 3 cycles/image of angular and radial frequency, respectively, and were selected based on previous
tests on a small-size dataset \cite{ZanaCesar2005}. We tested FBT descriptors of the whole image, or a combination of the upper right region, upper middle region, and the upper left region (Fig. \ref{fig:example}). In order to have a better notion of the information retained by the FBT, we used Eq. \ref{eq:ifbt} to reconstruct the image from the FB coefficients. The resulting image has a blurred aspect that reflects the use of only low-frequency radial components. In the rest of this paper, we will refer to the FB transformed images as just images. When using the PFT, the angular sampling was matched and only coefficients related to the same frequency range covered by the FBT were used. Both amplitude and phase information were considered.

\subsection{Polar frequency to dissimilarity domain}
We built a dissimilarity space $D\left( {{\rm {\bf t}},{\rm {\bf
t}}} \right)$ defined as the Euclidean distance between all
training FBT images ${\rm {\bf t}}$. In this space, each object is
represented by its dissimilarity to all objects. This approach is
based on the assumption that the dissimilarities of similar
objects to "other ones" is about the same \cite{Duinetal1997}.
Among other advantages of this representation space, by fixing the
number of features to the number of objects, it avoids a well
known phenomenon, where recognition performance is degraded as a
consequence of small number of training samples as compared to the
number of features.

\subsection{Classifier}
Test images were classified based on a pseudo Fisher linear
discriminant (FLD) using a two-class approach. A FLD is obtained by
maximizing the (between subjects variation)/(within subjects
variation) ratio. Here we used a minimum-square
error classifier implementation \cite{ScurichinaDuin1996}, which is
equivalent to the FLD for two-class problems. In
these cases, after shifting the data such that it has zero mean, the
FLD can be defined as
\begin{equation}
\label{eq:fld}
  g\left( {\bf x} \right)=\left[ {D\left( {{\rm
            {\bf t}},{\bf x}} \right)-\frac{1}{2}\left( {{\rm {\bf
              m}}_1 -{\rm {\bf m}}_2 }
      \right)} \right]^T{\rm {\bf S}}^{-1}\left( {{\rm {\bf m}}_1
      -{\rm {\bf m}}_2
    } \right)
\end{equation}
where ${\bf x}$ is a probe image, ${\rm {\bf S}}$ is the pooled
covariance matrix, and ${\rm {\bf m}}_i$ stands for the mean of
class $i$. The probe image ${\bf x}$ is classified as
corresponding to class-1 if $g({\bf x}) \ge 0$ and to class-2
otherwise.  However, as the number of objects and
dimensions is the same in the dissimilarity space, the sample
estimation of the covariance matrix ${\rm {\bf S}}$ becomes
singular and the classifier cannot be built. One solution to the
problem is to use a pseudo-inverse and augmented vectors
\cite{ScurichinaDuin1996}. Thus, Eq. \ref{eq:fld} is replaced by
\begin{equation}
  g\left( {\bf x} \right)=\left( {D\left( {{\rm
            {\bf t}},{\bf x}} \right),1} \right)\left( {D\left( {{\bf
            t},{\bf t}} \right),I} \right)^{\left( {-1} \right)}
 \end{equation}
 where $\left( {D\left( {{\rm {\bf t}},{\bf x}} \right),1} \right)$ is
 the augmented vector to be classified and $\left( {D\left( {{\bf
           t},{\bf t}} \right),I} \right)$ is the augmented training
 set. The inverse $\left( {D\left( {{\bf t},{\bf t}} \right),I}
 \right)^{\left( {-1} \right)}$ is the Moore-Penrose Pseudo-inverse
 which gives the minimum norm solution. The current $L$-classes
 problem can be reduced and solved by the two-classes solution
 described above. The training set was split into $L$ pairs of
 subsets, each pair consisting of one subset with images from a single
 subject and a second subset formed from all the other images.  A
 pseudo-FLD was built for each pair of subsets. A probe image was
 tested on all $L$ discriminant functions, and a ``posterior
 probability'' score was generated based on the inverse of the
 Euclidean distance to each subject.

\section{Database, preprocessing, and testing procedures}
The main advantages of the FERET database \cite{Phillipsetal1998} are the large number of individuals and rigid testing protocols that allow precise performance comparisons between different algorithms. We compare our algorithm performance with a "baseline" PCA-based algorithm \cite{TurkPentland1991} and with the results of three successful approaches. The PCA algorithm was based of a set of 700 randomly selected images from the gallery subset. The three first components, that are known to encode basically illumination variations, were excluded prior to image projecting. The other approaches are: Gabor wavelets combined with elastic graph matching \cite{Wiskottetal1997}, localized facial features extraction followed by a Linear Discriminant Analysis (LDA) \cite{EtemadChellappa1997}, and a Bayesian generalization of the LDA method \cite{Moghaddametal2000}. 

In the FERET protocol, a "gallery" set of one frontal view image from 1196 subjects is used to train the algorithm and a different dataset is used as probe. We used the probe sets termed "FB" and "DupI". These datasets contain single images from a different number of subjects (1195 and 722, respectively) with differences of facial expression and age, respectively. The "age" variation subset included several subjects that started or quited wearing glass or grow beards since their "gallery" pictures were takes.
Images were normalized using the eyes ground-truth information or coordinates given by an eyes detector algorithm. This face detection stage was implemented using a cascade of classifiers algorithm for the face detection \cite{ViolaJones2001} followed by an Active Appearance Model algorithm (AAM) \cite{Cootesetal2001} for the detection of the eyes region.  Within this region, we used flow field information \cite{KothariMitchell1996} to determine the eye center. Approximately 1\% of the faces were not localized by the AAM algorithm, in which cases the eyes regions coordinates were set to a fix value derived from the mean of the other faces. The final mean error was 3.7 $\pm$ 5.2 pixels. Images were translated, rotated, and scaled so that the eyes were registered at specific pixels (Fig. \ref{fig:example}). Next, the images were cropped to 130 x 150 pixels size and a mask was applied to remove most of the hair and background. The unmasked region was histogram equalized and normalized to mean zero and a unit standard deviation.

The system performance was evaluated by verification tests
according to the FERET protocol \cite{Phillipsetal1998}.  Given a
gallery image $g$ and a probe image $p$, the algorithm verifies
the claim that both were taken from the same subject. The
verification probability $P_{V}$ is the probability of the
algorithm accepting the claim when it is true, and the false-alarm
rate $P_{F}$ is the probability of incorrectly accepting a false
claim. The algorithm decision depends on the posterior probability
score $si\left( k \right)$ given to each match, and on a threshold
$c$. Thus, a claim is confirmed if $si\left( k \right)\le c$ and
rejected otherwise. A plot of all the combinations of $P_{V}$ and
$P_{F}$ as a function of $c$ is known as a receiver operating
characteristic (ROC).

\section{Results}
\begin{figure}[!t]
   \begin{center}
   \begin{tabular}{cc}

      \epsfig{file=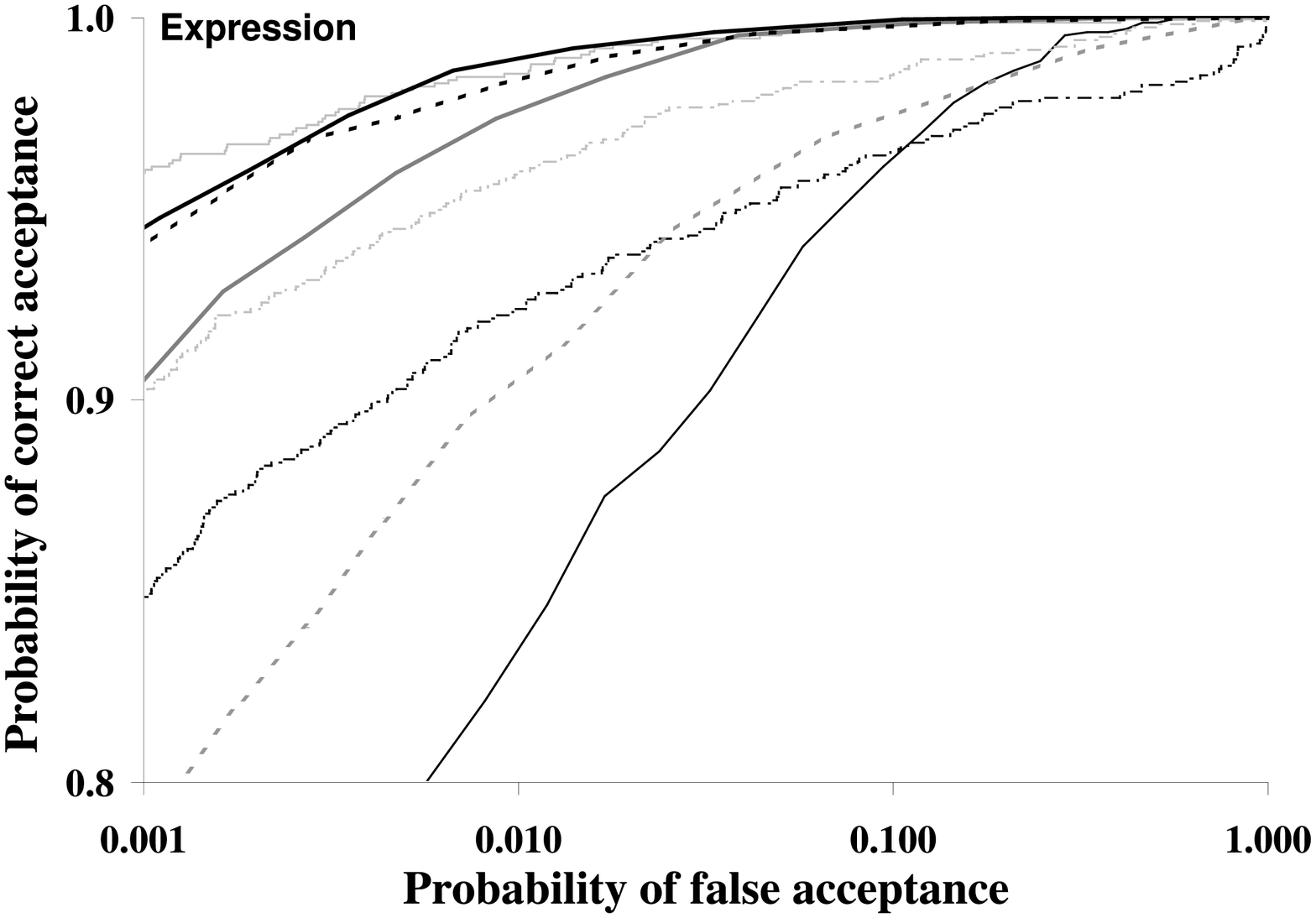, width=0.5\textwidth} &
      \epsfig{file=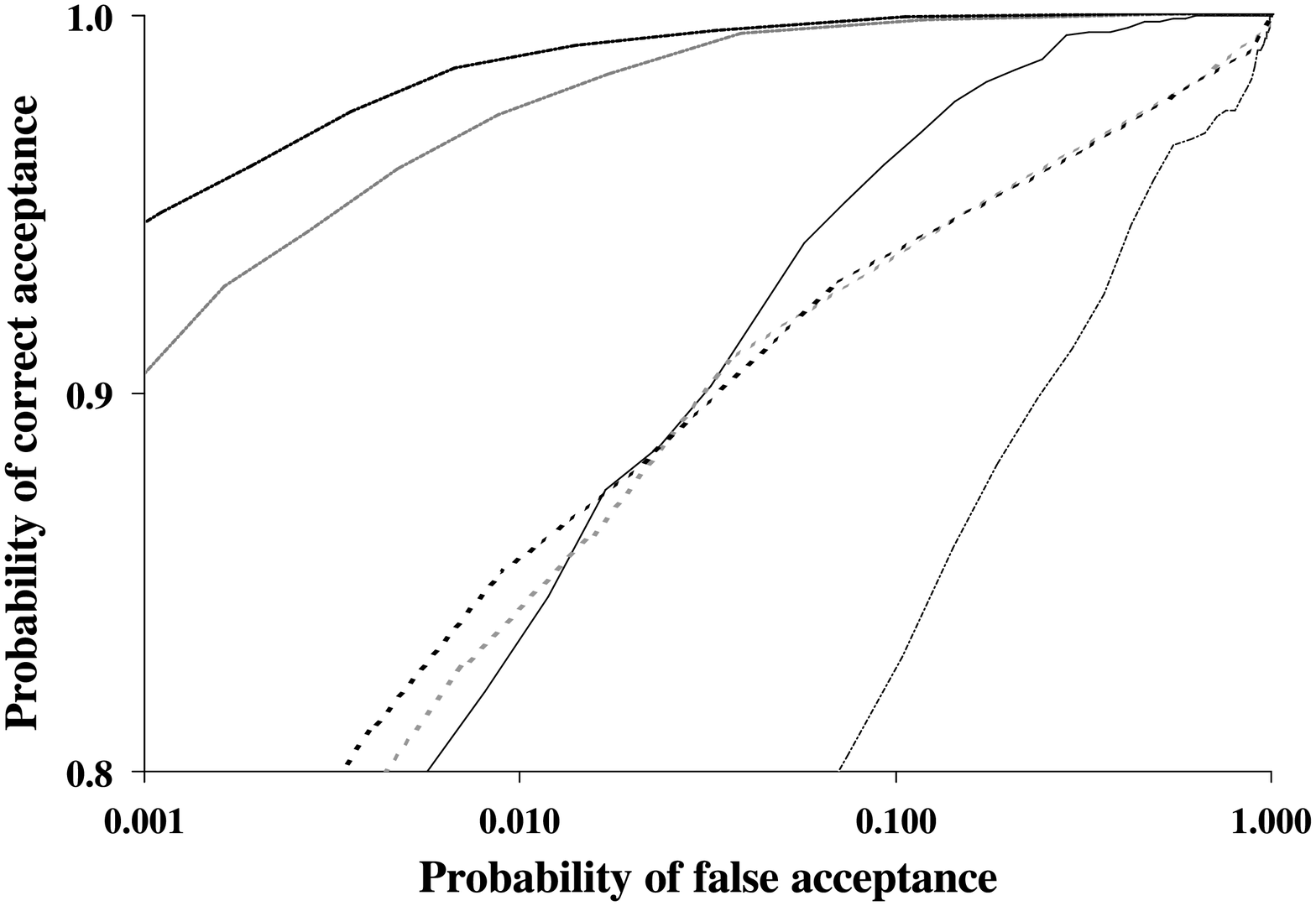, width=0.5\textwidth}
      \\
      \epsfig{file=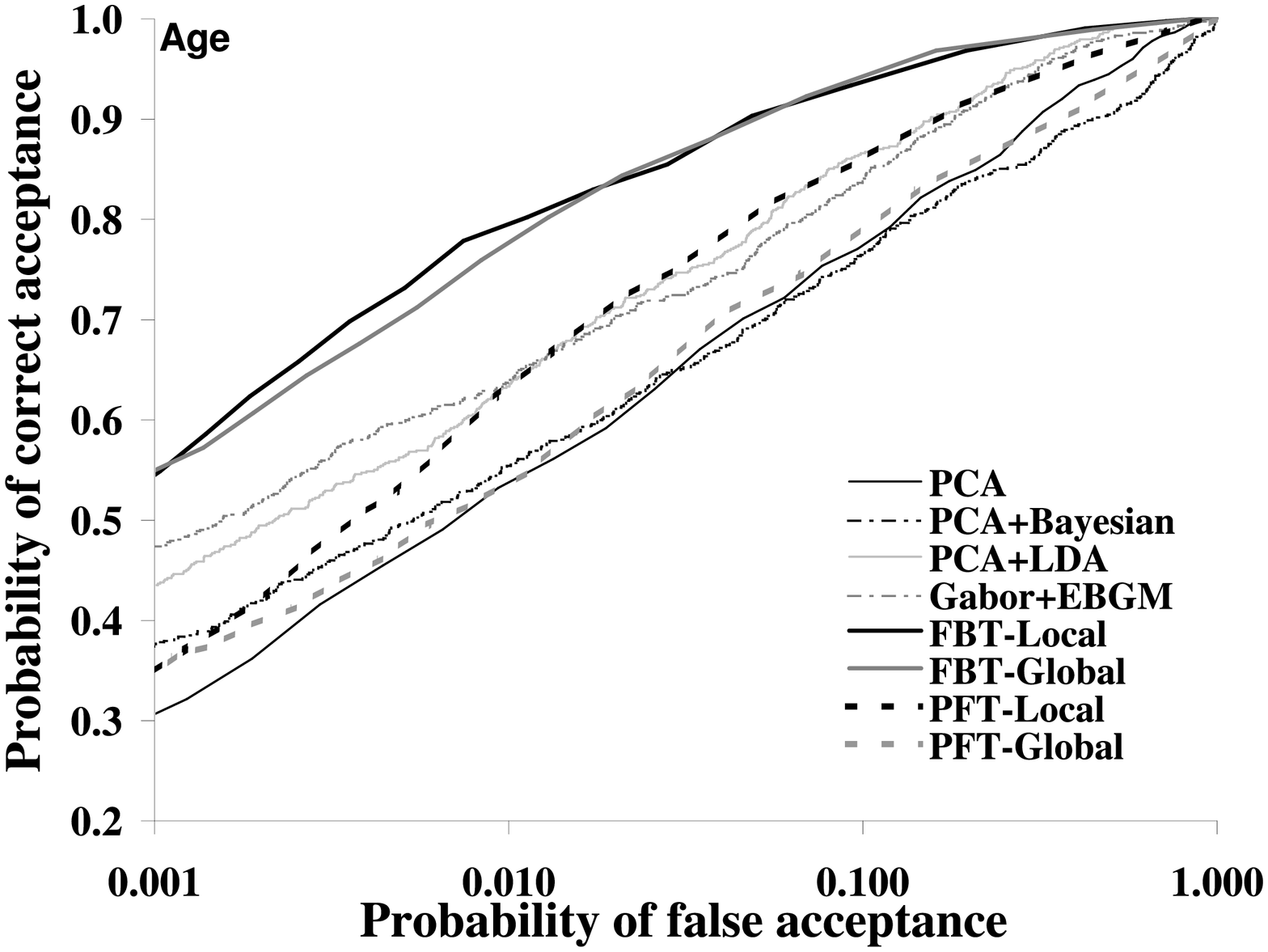, width=0.5\textwidth} &
      \epsfig{file=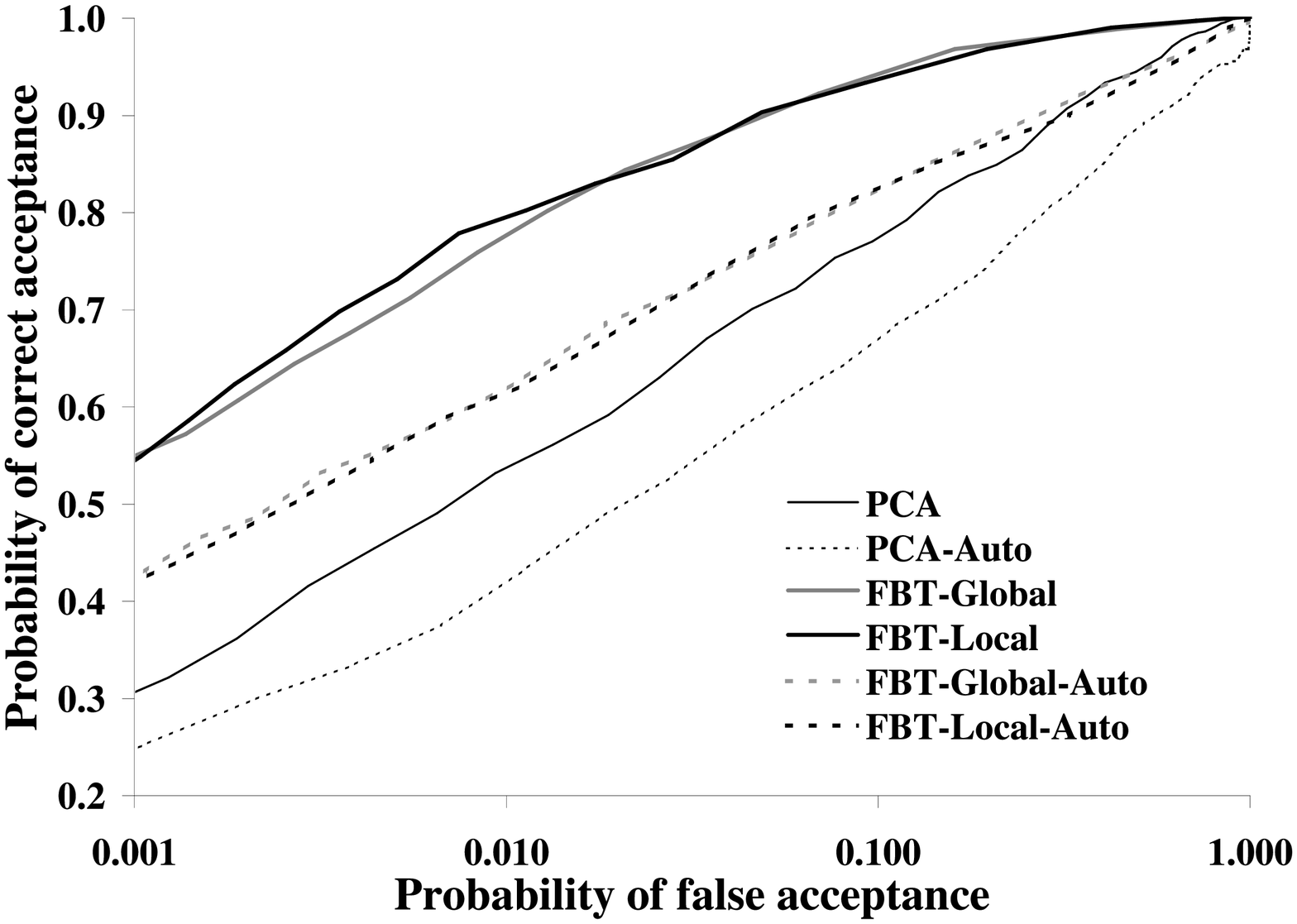, width=0.5\textwidth}
   
   \end{tabular}
   \caption{ROC functions of the proposed and previous algorithms. Left panels: using semi-automatic FBT and PFT. Right panels: Using automatic FBT}
   \label{fig:roc}
   \end{center}
\end{figure}

Figure \ref{fig:roc} shows the performance of the proposed verification system. The local FBT version performed at the same level of the best previous algorithm (PCA+LDA) on the expression dataset and achieved the best results on the age subset. The global version of the FBT algorithm was inferior at all conditions. Comparisons with the PFT representation indicate that this alternative features are less robust to age and illumination variations. Automation of the eye detection stage reduced the system performance by up to 20\%. This reduction is expected, considering the variance property of the FBT to translation \cite{Cabreraetal1992}, and reflect sensitivity to registration errors typical to other algorithms, like the PCA. 

We also computed the equal error rate (EER) of the proposed algorithms (Table \ref{table:eer}). The EER occurs at a threshold level where the incorrect rejection and false alarm rates are equals (1-$P_{V}=P_{F}$). Lower values indicate better performance. The EER results reinforce the conclusions from the ROC functions.

\begin{table}
  \begin{center}
    \caption{Equal error rate (\%) of the FBT, PFT and previous algorithms}
  \begin{tabular}{l|cccc}

       &
       \multicolumn{2}{c}{\bf Semi-Auto} & 
       \multicolumn{2}{c}{\bf Auto} \\
       \hline
  
      \bf Algorithm &
      \bf Expression &
      \bf Age &
      \bf Expression &
      \bf Age \\
      \hline

      \bf FBT-Global &
      1.6 &
      7 &
      4.5 &
      16 \\

      \bf FBT-Local &
      1.1 &
      8 &
      7.1 &
      16 \\

      \bf PFT-Global &
      4.1 &
      16 \\

      \bf PFT-Local &
      1.4 &
      12 \\

      \\

      PCA &
      5.9 &
      17 &
      14 &
      23 \\

      PCA+Bayesian &
      4.9 &
      18 \\

      PCA+LDA &
      1.2 &
      13 \\

      Gabor-EBGM &
      2.5 &
      13 \\

  \end{tabular}
  \label{table:eer}
  \end{center}
\end{table}

\section{Discussion}

Most of the current biologically-inspired algorithms were validated only on very small databases, with the exception of the Gabor-EBGM algorithm that is based on Gabor wavelets transformation and can be viewed as analogous to human spatial filtering at the early stages. Here we presented a novel face verification system based on a local analysis approach and FBT descriptors. The system achieving top ranking on both the age and expression subset. These results are a clear demonstration of the system robustness in handling realistic situations of facial expression and age variations of faces.

The local approach is also superior to the global, a fact that have direct implication for the algorithm robustness for occlusions. In the local approach, the mouth region is ignored, thus its occlusion or variation (ex. due to a new beard) does not affect performance at all. From the computational point of view, the local analysis does not imply much more computation time: the FBT of each region consumes about half the time consumed by the global analysis. Preliminary results (not shown) of the global version with reduced resolution images indicate that computation time can be further with no performance loss, but we still have not tested the effect of image resolution on the local version.

The system have an automatic face detection version, but the trade-off is a certain performance loss. We currently work on the implementation of more precise face and eye detectors algorithms. For example, \cite{Martinez2002} learned the subspace that represents localization errors within eigenfaces. This method can be easily adopted for the FBT subspace, with the advantage of the option to exclude from the final classification face regions that gives high localization errors. 

Currently, we are developing a series of psychophysical experiments with the aim of establishing the relation of the proposed system with human performance. The main questions are: (1) At what location and scale global spatial pooling occurs? (2) Are faces represented in a dissimilarity space? (3) How does filtering of specific polar frequency components affects the face recognition performance of humans and the proposed system?

In conclusion, the proposed system achieved state-of-the-art performance in handling problems of expression and age variations. We expect from future tests to show robustness to illumination variation and partial occlusion and our on-going work are focused on improving the performance of the automatic version.

\bibliographystyle{splncs}
\bibliography{bib_yossi_050413}

\end{document}